\documentclass[10pt,twocolumn,letterpaper]{article}

\usepackage{iccv}
\usepackage{times}
\usepackage{epsfig}
\usepackage{graphicx}
\usepackage{amsmath}
\usepackage{amssymb}

\usepackage{subfigure}

\usepackage[pagebackref=true,breaklinks=true,letterpaper=true,colorlinks,bookmarks=false]{hyperref}

\iccvfinalcopy 

\def\iccvPaperID{9} 

\ificcvfinal\fi
\begin{document}

\title{Accurate and Efficient Similarity Search for Large Scale Face Recognition}

\author{Ce Qi\\
BUPT\\
\\
{\tt\small }
\and
Zhizhong Liu\\
BUPT\\
\\
{\tt\small }
\and
Fei Su\\
BUPT\\
\\
{\tt\small }
}

\maketitle

\begin{abstract}

Face verification is a relatively easy task with the help of discriminative features from deep neural networks. However, it is still a challenge to recognize faces on millions of identities while keeping high performance and efficiency. The challenge 2 of MS-Celeb-1M\cite{guo2016msceleb,facelowshot} is a classification task. However, the number of identities is too large and it is not that elegant to treat the task as an image classification task. We treat the classification task as similarity search and do experiments on different similarity search strategies. Similarity search strategy accelerates the speed of searching and boosts the accuracy of final results. The model used for extracting features is a single deep neural network pretrained on CASIA-Webface\cite{yi2014learning}, which is not trained on the base set or novel set offered by official. Finally, we rank \textbf{3rd}\footnote{Leaderboard: https://www.msceleb.org/leaderboard/c2} in challenge 2 of MS-Celeb-1M\cite{guo2016msceleb,facelowshot}, while the speed of searching is 1ms/image.

\end{abstract}

\section{Introduction}

Recently, deep neural networks\cite{krizhevsky2012imagenet,lecun1989backpropagation} have achieved state of the art performance in computer vision tasks, such as visual object classification and recoginition\cite{krizhevsky2012imagenet,he2015deep,xie2016aggregated}, showing the power of the extracted discriminative features. The performance of face recognition\cite{sun2013hybrid,taigman2014deepface,sun2014deep_0,sun2014deep_1,sun2015deeply,schroff2015facenet,wen2016discriminative} has been boosted for the discriminative features extracted from deep neural networks, especially CNN\cite{krizhevsky2012imagenet, he2015deep, xie2016aggregated}, which mainly benefits from the large scale of training data\cite{schroff2015facenet, krizhevsky2012imagenet} and computing resources.

The challenge 2 of MS-Celeb-1M\cite{guo2016msceleb,facelowshot} is a problem of low-shot face recognition, with the goal to build a large-scale face recognizer capable of recognizing a substantial number of individuals with high precision and high recall. The goal is to study when tens of images are given for each person in the base set while only one to five images are given for each person in the novel set. The main difficulty of challenge 2 is: (a) the small amount of training data for novel set; (b) the number of training data for base set is much large than novel set's. Therefor, we develop an algorithm to recognize the persons in both the data sets, by proposing a new pipeline for face recognition/search.

\subsection{Face recognition}

For face recognition task, there are several works learning with even more discriminative features to further improve the performance of deep neural networks. Some researchers use deeper, wider and more complex network structures to obtain better features\cite{he2015deep,xie2016aggregated}. There are also some other efforts on new non-linear activations\cite{he2015delving,goodfellow2013maxout}, dropout\cite{krizhevsky2012imagenet} and batch normalization\cite{ioffe2015batch} to make the networks perform better.

Except for powerful network structure, another kind of approaches to improve the performance of face recognition is metric learning. Using auxiliary loss to supervise the training of networks is a simply way in metric learning, such as contrastive loss\cite{sun2014deep_1}, triplet loss\cite{schroff2015facenet}, center loss\cite{wen2016discriminative}, contrastive-center loss\cite{qi2017contrastive-center} and NormFace\cite{wang2017normface}. These approaches are proposed for the purpose of more discriminative features by enforcing better intra-class compactness and inter-class separability. The contrastive loss\cite{sun2014deep_1} and triplet loss\cite{schroff2015facenet} do really improve the performance of networks but they all need carefully selected pairs or triplets. And the selection of pairs and triplets have influences on the training results of deep neural networks. What's more, if all possible training samples combinations are chosen, the number of training pairs and triplets would theoretically go up to O($N^2$), where N is the total number of training samples. The center loss\cite{wen2016discriminative} and contrastive-center loss\cite{wen2016discriminative}, which learn a center for each class, both are a kind of new novel loss to enforce extra intra-class compactness or inter-class separability. The center loss does not consider the inter-class separability, but contrastive-center loss does. The NormFace\cite{wang2017normface} considers feature normalization and optimizes cosine similarity directly instead of inner-product.

However, it is still a challenge to do face recognition in the wild, mainly because of the large variance of faces. In case of identifying one face in more than one million faces, conventional approach just compares the cosine similarity simply, which may decrease the performance and not that efficient.

\subsection{Similarity search}

Nowadays, there are large amounts of images and videos, especially from the Internet. How to get the images or videos we are interested in are important. However, interpretation and searching of images and videos are not that easy and require accurate and efficient algorithms. A variety of machine learning and deep learning algorithms are being used to help the interpretation and searching of these complex, real-world entities.

In this context, searching by numerical similarity rather via structured relations is more suitable\cite{johnson2017billion,gordo2016deep}. Using the similarity could find the most similar content to a picture, or find the vectors that are most similar.

State of the art similarity search methods like NN-Descent\cite{dong2011efficient} have a large memory overhead on top of the dataset itself and cannot readily scale to billion-sized databases, such as MS-Celeb-1M\cite{guo2016msceleb,facelowshot}. Therefore, using NN-Desent like methods to compute a \emph{k}-NN graph is not practical. Rendering both exhaustive search and exact indexing for non-exhaustive search are impractical on billion-sized databases. So, the approximate search or graph construction is fit for the task of efficient and accurate similarity search on billion-sized datasets.

The methods based on product quantization(PQ) codes\cite{jegou2011product,johnson2017billion,babenko2012inverted,ge2014optimized} are shown to be more effective than binary codes\cite{norouzi2013cartesian}, for the binary codes incur important overheads for non-exhaustive search\cite{norouzi2012fast}. However, most of the methods about PQ are difficult to implement efficiently on GPUs\cite{johnson2017billion}.

There are also many other implementations of similarity search on GPUs\cite{johnson2017billion}, but most of them are with binary codes\cite{pan2011fast}, small datasets\cite{wakatani2014gpgpu}, or exhaustive search\cite{dashti2013efficient,sismanis2012parallel,tang2015efficient}, which are not elegant.

The most outstanding similarity search methods based on PQ are \cite{wieschollek2016efficient} and \cite{johnson2017billion}. Using effective similarity search method will help the task of similarity search significantly.

\subsection{Combining similarity search with face recognition}

To boost the performance of face recognition, there are many strategies, such as using very large scale training data\cite{schroff2015facenet}, metric learnings\cite{sun2014deep_1,schroff2015facenet,wen2016discriminative,qi2017contrastive-center,wang2017normface} and deeper and wider neural networks\cite{he2015deep,xie2016aggregated}. However, these strategies are not that appropriate in the condition of face recognition 1:N when N is a too large number. In some cases, the base search datasets may contain some error-labeled face images or non-face images. Otherwise, the internal structure and contact of base search datasets is a good clustering reference, which is significant for face recognition in billion-sized datasets. In the condition of searching faces when base dataset is billion-sized, it is necessary to adopt some strategies to speed up the searching while keeping high accuracy.

In face verification or face recognition, similarity score is mostly used to indicate the similarity of two faces. To speed up the searching while keeping high accuracy in million-sized face recognition, such as MS-Celeb-1M\cite{guo2016msceleb,facelowshot}, we adopt similarity search strategies based on product quantization\cite{johnson2017billion}, which are most efficient and accurate similarity search methods. What's more, these methods are implemented on GPUs, which will boost the search speed further.

\begin{figure}
\centering
\subfigure[preprocessing]
{
\label{fig:preprocessing}
\includegraphics[width=0.45\textwidth]{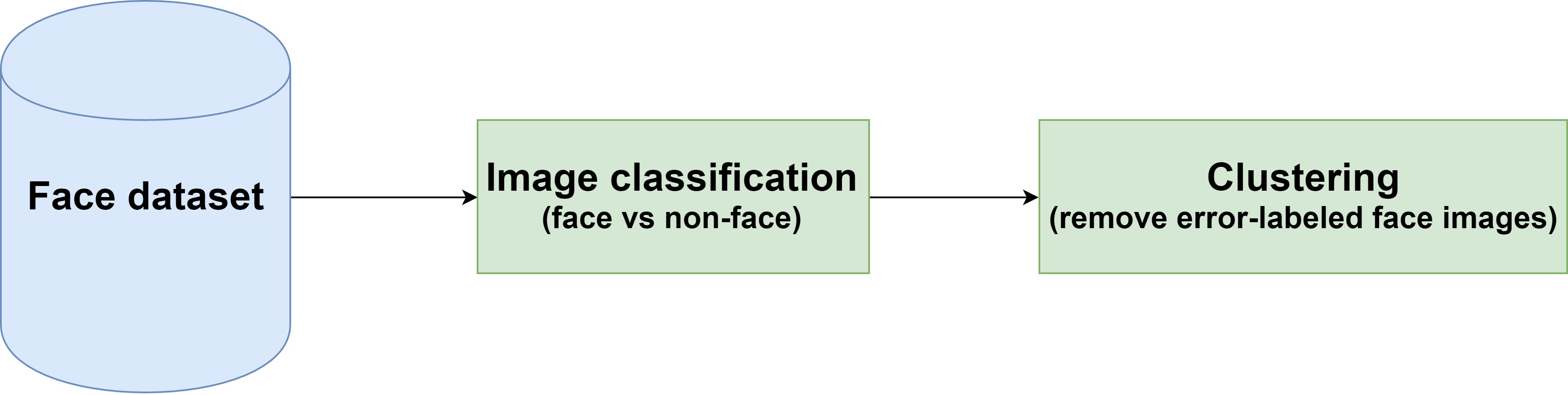}
}
\hspace{0in}
\subfigure[face search]
{
\label{fig:face search}
\includegraphics[width=0.45\textwidth]{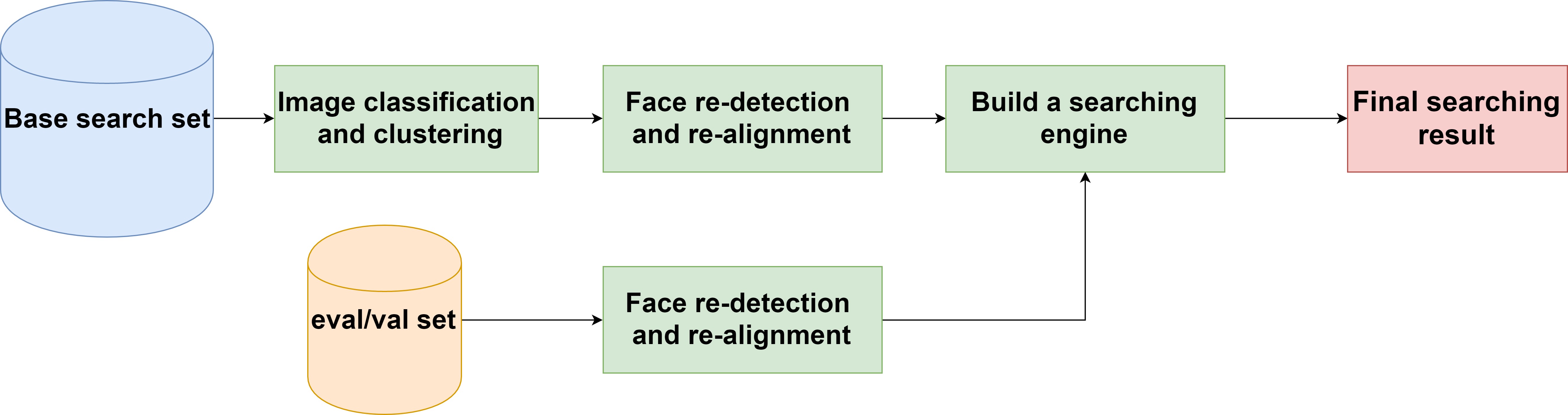}
}
\caption{Face search pipeline}\label{fig:face search pipeline}
\end{figure}

In this paper, we will propose a new pipeline for face recognition/search on very large scale face datasets, as shown in Fig. \ref{fig:face search pipeline}. It mainly consists of two stages:

1. In the stage of preprocessing, image classification and clustering are used to remove the non-face images and error-labeled images.

2. In the stage of searching, face re-detection and re-alignment are used to get better normalized face images. A pretrained deep neural network followed to extract features. Finally, PQ methods are introduced to establish a search engine for efficient and accurate face search. And the efficient implementation on GPUs will boost the speed of searching further.

The details will be described in the later sections.

\section{Preprocessing on face images}

\begin{figure}
\centering
\includegraphics[width=0.3\textwidth]{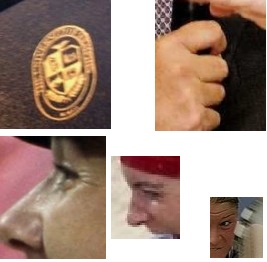}
\caption{Non-face images and partial face images. These images will decrease the accuracy of face recognition.}\label{fig:non-face-by-image-classification}
\end{figure}

\begin{figure}
\centering
\includegraphics[width=0.5\textwidth]{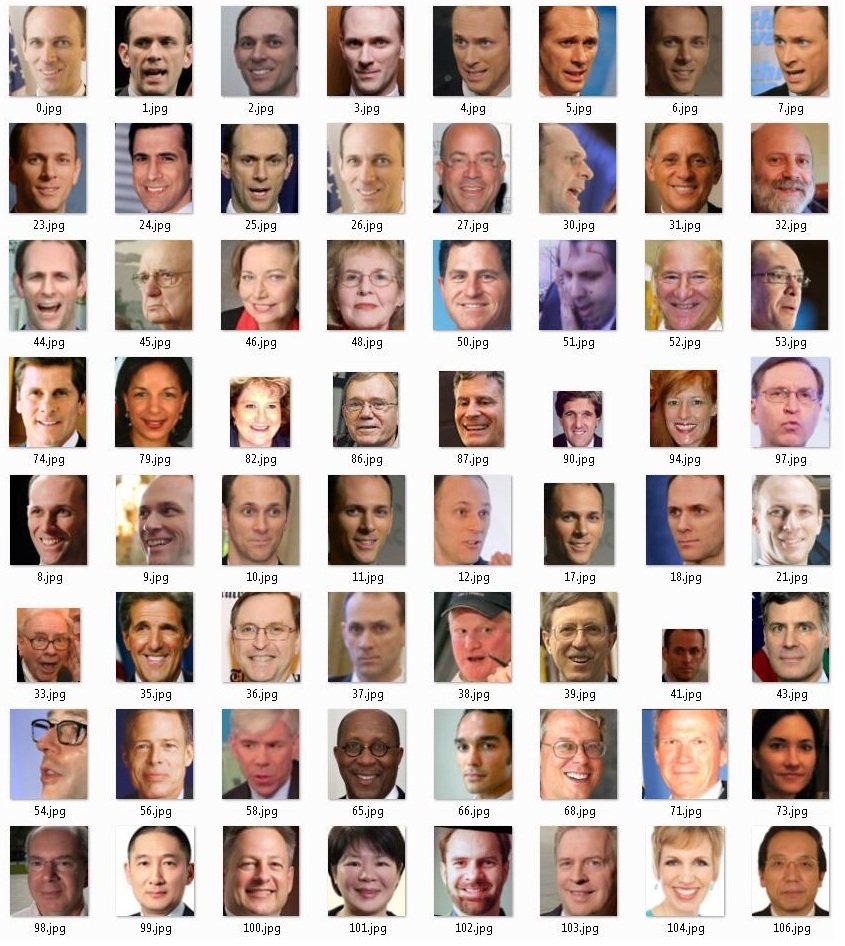}
\caption{Error-labeled face images. These images are in one folder. All images in this folder have same label, but belong to different people.}\label{fig:error-labeled_face_img}
\end{figure}

 There are many non-face images and error-labeled face images, as shown in Fig. \ref{fig:non-face-by-image-classification} and Fig. \ref{fig:error-labeled_face_img}, which will decrease the final accuracy. Therefore, image classification and clustering are used firstly in our proposed scheme to remove the error-labeled face images and non-face images in the large scale face datasets\cite{guo2016msceleb,facelowshot}. Then, face re-detection and re-alignment is used to improve the accuracy of face recognition further.

\subsection{Image classification and clustering}\label{sec:image classfication and clustering}

Firstly, we train an image classification model to class face and non-face. The dataset used is WIDER FACE\cite{yang2016wider}. The detail of training the network is described in Section \ref{sec:experiments on preprocessing}. After training the classification network, all the images in base and novel set in challenge 2 of MS-Celeb-1M\cite{guo2016msceleb,facelowshot} are classified by the network and only the face images suitable for face recognition are kept. In the process of classification, face images not containing full faces are removed.

Secondly, to remove those error-labeled face images, clustering is done for every image folders in the base set. Face images of different people are divided into corresponding folders, which are named by Freebase MID. One folder should contains images of one person. Clustering on every folder is proposed through features extracted from a deep neural network. The number of clustering center of every folder is set to 2, and the center has more points will be set as main center. The scheme of removing error-labeled face images is to compare the extracted features' distance between the image and main center. If the distance is larger than twice of the average distance of corresponding points, the image will be removed. After the clustering, most of the folders in the base set only contain single person's face images.

\subsection{Face re-detection and re-alignment}\label{sec:face re-detection and re-alignment}

\begin{figure}
\centering
\includegraphics[width=0.5\textwidth]{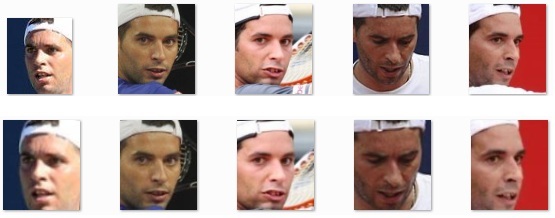}
\caption{Original face images VS normalized face images. Do re-normalization will make the accuracy of face recognition increased, which can be seen when comparing Table \ref{table:using original images} and Table \ref{table:using images processed by face re-detection and re-alignment}}\label{fig:ori_vs_norm}
\end{figure}

Although the dataset have been cleaned and normalized, some images are not well aligned, such as the first row of Fig. \ref{fig:ori_vs_norm}. What is more, to improve the accuracy of face recognition, using same face detection and alignment methods to do face model training and testing is very significant, which can be seen when comparing Table \ref{table:using original images} and Table \ref{table:using images processed by face re-detection and re-alignment}. Therefore, after removing non-face images and error-labeled face images, we do face re-detection and re-alignment.

\begin{figure}
\centering
\includegraphics[width=0.3\textwidth]{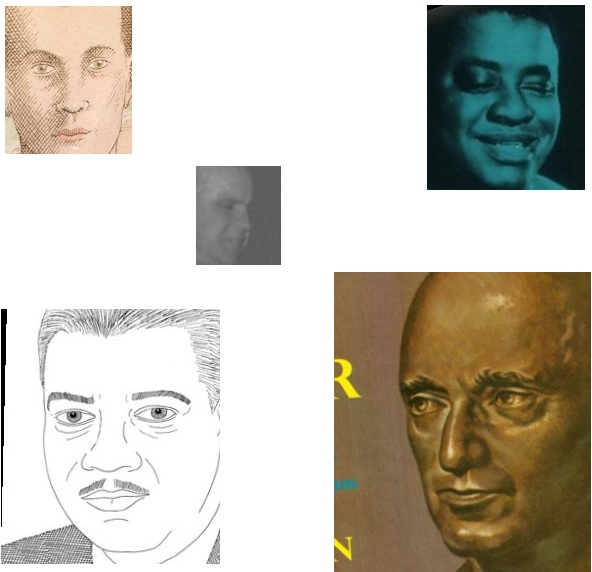}
\caption{Some examples of face images hard to do face detection}\label{fig:hard_detect_img}
\end{figure}

\begin{figure}
\centering
\subfigure[a]
{
\label{fig:preprocessing}
\includegraphics[width=0.1\textwidth]{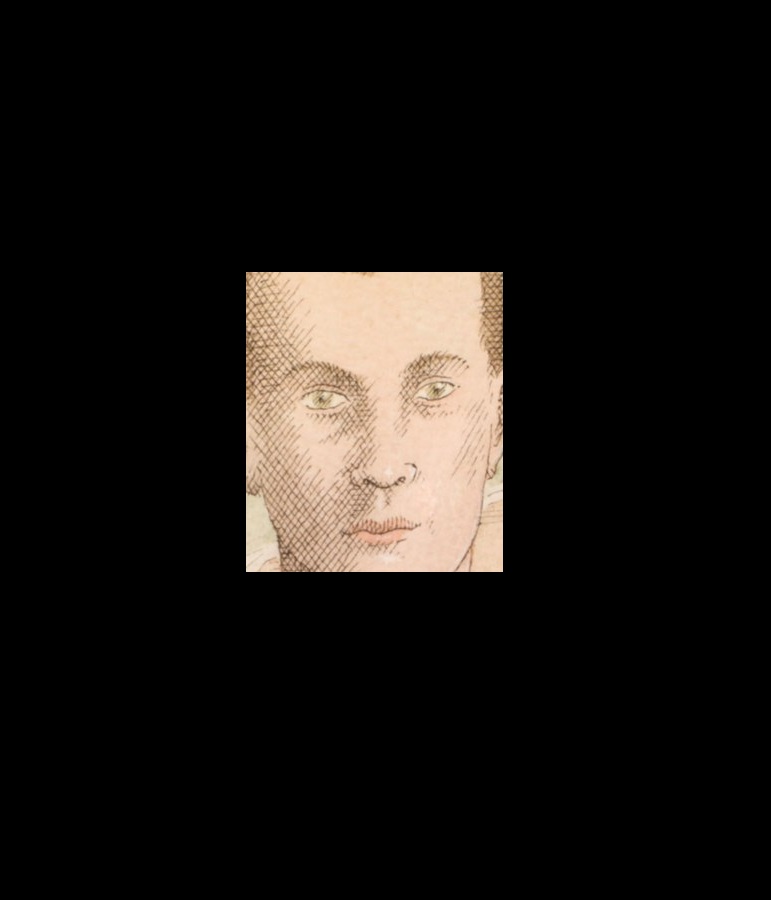}
}
\hspace{0in}
\subfigure[b]
{
\label{fig:face search}
\includegraphics[width=0.1\textwidth]{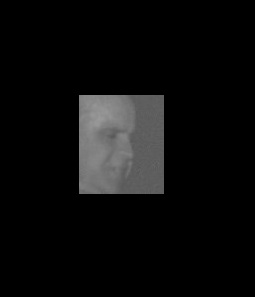}
}
\hspace{0in}
\subfigure[c]
{
\label{fig:face search}
\includegraphics[width=0.1\textwidth]{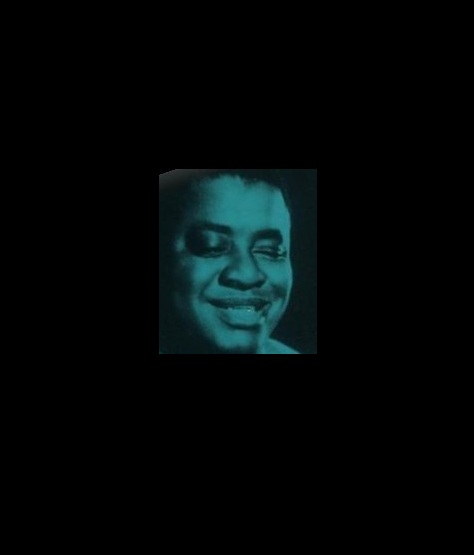}
}
\hspace{0in}
\subfigure[d]
{
\label{fig:face search}
\includegraphics[width=0.1\textwidth]{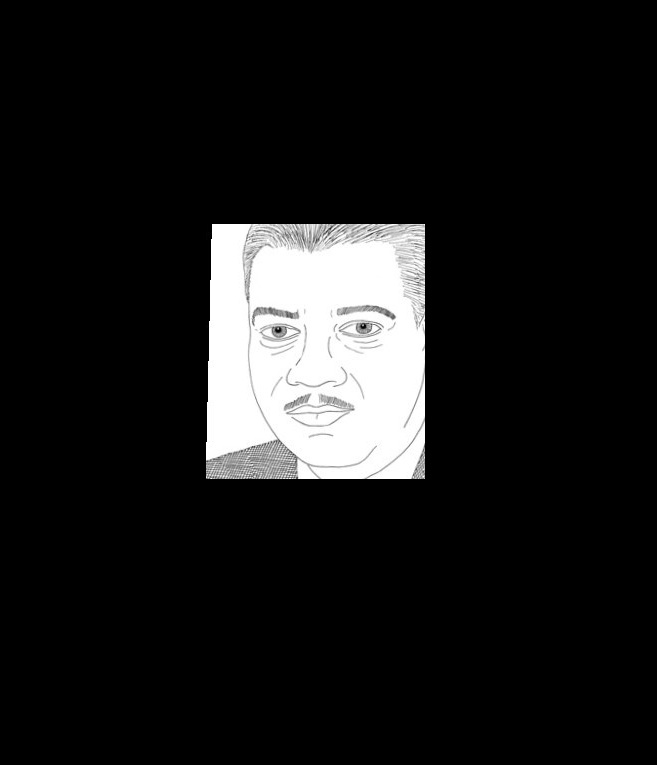}
}
\caption{Face image padding}\label{fig:padding for face iamges hard to detect}
\end{figure}

Note that many faces in the dataset\cite{guo2016msceleb,facelowshot} are hard to be detected. As shown in Fig. \ref{fig:hard_detect_img}, the face detector cannot detect the faces, for the face is too large or the quality of images are not that good. So, we use some tricks to handle this problem. As shown in Fig. \ref{fig:padding for face iamges hard to detect}, we pad zeros for those face images hard to be detected. The padding size in width and height is equal to the corresponding width and height of images. The padding will make the faces are relatively easier to be detected. If faces cannot be detected after padding, the original images will be used.

\section{Discriminative face feature extractor}

A deep neural network is used as feature extractor. And the input images are normalized by face detection and alignment through similarity transformation. Note that the model for extracting features is only trained on CASIA-Webface\cite{yi2014learning}, not trained on MS-Celeb-1M\cite{guo2016msceleb,facelowshot}.

\subsection{Face detection and alignment}\label{sec:face detection and alignment}

Face features can be extracted through a deep neural network. Before extracting features, face detection and alignment will be done here. MTCNN\cite{7553523} algorithm is used for face detection and landmark detection. We use 5 landmarks (two eyes, nose and
mouth corners) for similarity transformation. The face images are cropped to $112\times96$ RGB images, and each pixel (in [0,255]) in RGB images is normalized by subtracting 127.5 then divided by 128.

\subsection{Face recognition model}

After preprocessing, face detection and alignment, we train a single deep neural network for feature extraction, called FRN(or FaceResNet), which is the same as NormFace\cite{wang2017normface}. The structure of the deep neural network is shown in Table \ref{table:FRN}. The FRN is trained under the supervision of improved softmax loss used in NormFace\cite{wang2017normface} and contrastive-center loss\cite{qi2017contrastive-center} jointly to get more discriminative feature.


\begin{table}
\begin{center}
\scalebox{0.6}
{
\begin{tabular}{|c|c|}
\hline
layer name&FRN\\
\hline
\hline
conv1a&$3\times3$, 32\\
\hline
conv1b&$3\times3$, 64\\
\hline
pool1b&$2\times2$, max pool, stride 2\\
\hline
conv2\_x&$1\times\begin{cases}3\times3, 64\\ 3\times3, 64\end{cases}$\\
\hline
conv2&$3\times3$, 128\\
\hline
pool2&$2\times2$, max pool, stride 2\\
\hline
conv3\_x&$2\times\begin{cases}3\times3, 128\\ 3\times3, 128\end{cases}$\\
\hline
conv3&$3\times3$, 256\\
\hline
pool3&$2\times2$, max pool, stride 2\\
\hline
conv4\_x&$5\times\begin{cases}3\times3, 256\\ 3\times3, 256\end{cases}$\\
\hline
conv4&$3\times3$, 512\\
\hline
pool4&$2\times2$, max pool, stride 2\\
\hline
conv5\_x&$3\times\begin{cases}3\times3, 512\\ 3\times3, 512\end{cases}$\\
\hline
fc5&512\\
\hline
Loss&improved softmax loss\cite{wang2017normface} and contrastive-center loss\cite{qi2017contrastive-center}\\
\hline
\end{tabular}
}
\end{center}
\caption{The CNNs architecture\cite{wang2017normface} use for face feature extractor, which is called FRN(or FaceResNet). "$3\times3$, 64" denotes convolution layers with $64$ filters of size $3\times3$. "stride 2" denotes the stride size is 2. "max pool" denotes max pooling. The brace and its left number denote the structure will duplicate for times of the left number.}\label{table:FRN}
\end{table}

\section{Similarity search for face recognition}

After the preprocessing of cleaning non-face images and error-labeled face images of challenge 2 in MS-Celeb-1M\cite{guo2016msceleb,facelowshot} and the training of face feature extractor, we implement an accurate and efficient face recognition/search engine using similarity search methods based on PQ. Here, the open-source framework Faiss\cite{Faiss}, based on \cite{johnson2017billion}, is used.

The face search engine construction step is as follows:

1. Extracting features of cleaned base set and top-1 novel set of challenge 2 in MS-Celeb-1M\cite{guo2016msceleb,facelowshot} as base search set.

2. Using different strategies for face search, and testing on the validation set developed by ourselves and development set offered by official\cite{guo2016msceleb,facelowshot}.

\section{Experiments}\label{sec:experiments}

\begin{figure}
\centering
\hspace{0in}
\subfigure[]
{
\label{fig:list_diy_validation_set}
\includegraphics[width=0.3\textwidth]{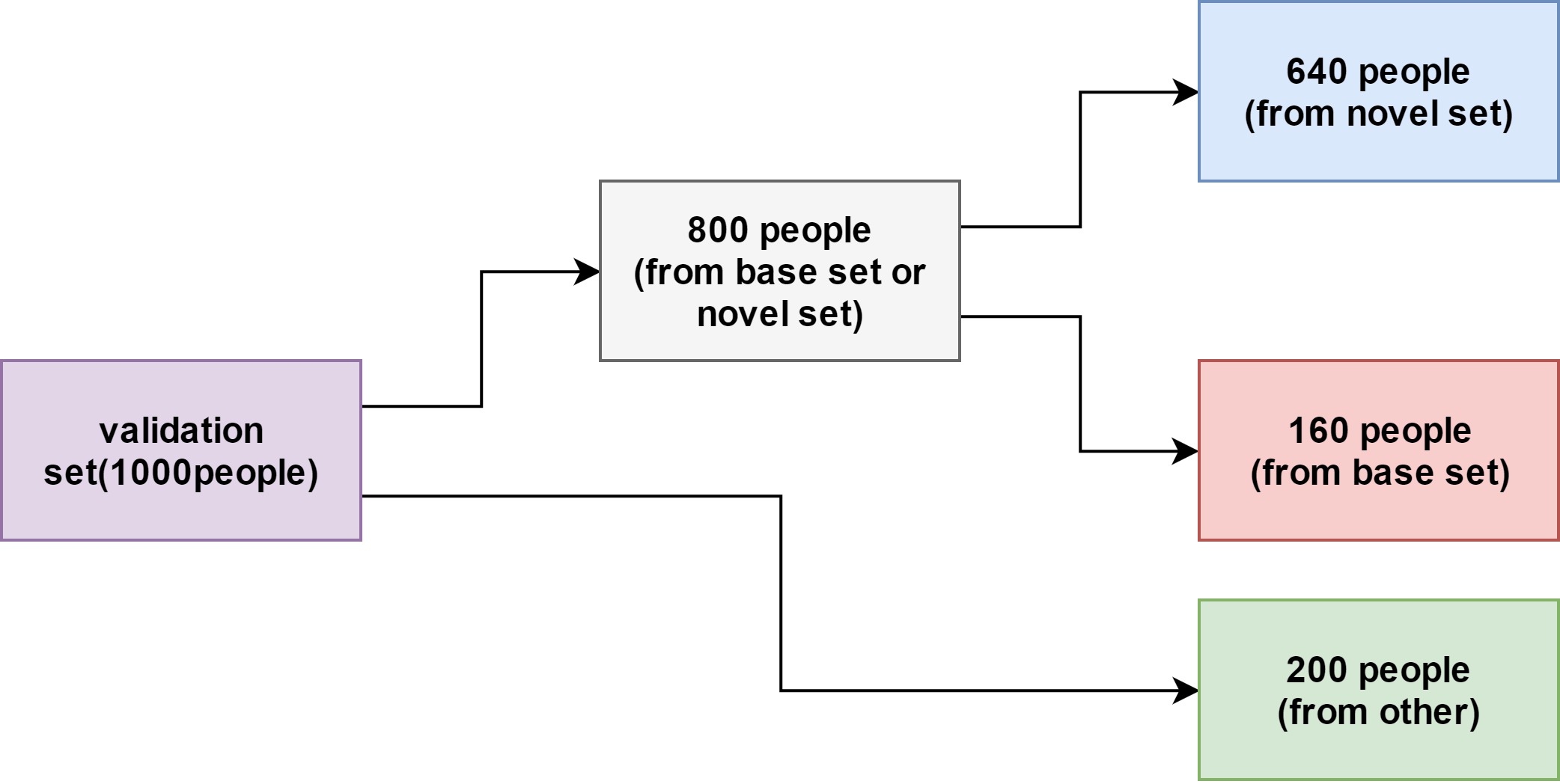}
}
\hspace{0in}
\subfigure[]
{
\label{fig:sector_diy_validation_set_whitebg}
\includegraphics[width=0.3\textwidth]{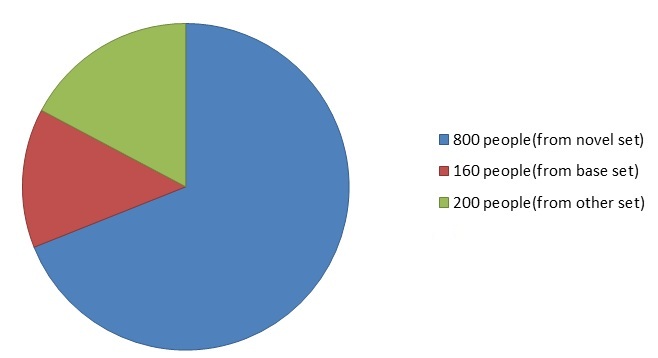}
}
\caption{Validation set made by ourself.}\label{fig:validation set made by ourself}
\end{figure}

To evaluate the performance of similarity search engine for face recognition/search, we make validation set by ourself. The validation set contains images of 1,000 people, while 800(80\%) people from base set or novel set of challenge 2 and 200(20\%) from other set(individuals belongs to MS-Celeb-1M but not in base set or novel set in challenge 2). And Fig. \ref{fig:validation set made by ourself} shows the detail of the validation set made by ourself. We randomly select three face images for each 1,000 people, so the total number of face images in the validation set is 3,000. Note that the images chosen as validation set are independent of the images used as base search set, which consist of images from base set and novel set.

And the development set offered by official\cite{guo2016msceleb,facelowshot} consists of 20,000 face images from base set and 5,000 face images from novel set, totally 25,000 face images.


\subsection{Experiments on preprocessing}\label{sec:experiments on preprocessing}

As mentioned before, the image classification and clustering are used to remove non-face images and error-labeled face images.

Caffenet\cite{jia2014caffe}, which is pretrained on ImageNet dataset\cite{krizhevsky2012imagenet}, is used as the image classification model. The size of input image is 256, then cropped to 224. For the training set and validation set, random crop is adopted on the images to sample positive samples and negative samples. If a crop has overlap higher than 0.85 with any ground-truth box, it will be treated as a positive sample. Otherwise, if it is lower than 0.7 with any ground-truth box, it will be treated as a negative sample. we empirically set the boundary, which makes the positive samples are all full face images. Note that all the positive and negative samples are augmented only by flipping and dimming. Other augmentation strategies are not suitable for the network to classify whether one image is appropriate for face recognition.

The final accuracy on validation set is nearly 99.5\%, which proves that the image classification is robust enough to remove non-face images and partial face images.



\subsection{Experiments on LFW}\label{sec:experiments on LFW}

MTCNN algorithm\cite{7553523} is used to do face detection and alignment, as described in section \ref{sec:experiments on LFW}.

The training data is cleaned for wrong collected images. People overlapping between the outside training data and the LFW testing data are excluded. After getting all normalized face images of CASIA-Webface\cite{yi2014learning}, we train the network under the supervision of improved softmax loss used in NormFace\cite{wang2017normface} and contrastive-center loss\cite{qi2017contrastive-center} jointly. PCA is not used for face recognition here. When testing, original image and its flipped one are used to extract features. Then, two 512-d features are fused into one 512-d by eltwise sum. Table \ref{table:accuracy of different feature fusing strategy} shows the details of different feature fusing strategies. The best final accuracy on LFW is $99\%$ with eltwise strategy of sum.

\begin{table}
\begin{center}
\scalebox{0.6}
{
\begin{tabular}{|c|c|}
\hline
feature fusing strategy&accuracy on LFW(\%)\\
\hline
\hline
single image&98.8\\
\hline
concatenate&98.82\\
\hline
SORT&98.83\\
\hline
PROD&98.1\\
\hline
\textbf{SUM}&\textbf{99}\\
\hline
MAX&98.97\\
\hline
\end{tabular}
}
\end{center}
\caption{Accuracy of different feature fusing strategy on LFW\cite{LFWTech}. The accuracy of eltwise sum is the highest.}\label{table:accuracy of different feature fusing strategy}
\end{table}


\subsection{Experiments on search strategies}








To evaluate the performance of the proposed pipeline for face recognition/search on large scale datasets, comparisons of different search strategies are given on validation set and development set.

The processed base set and top-1 novel set are used as base search set, and the images chosen as validation set are excluded from the base search set in the stage of testing.

The processing of image classification and clustering to remove non-face images and error-labeled face images are excluded in testing.

Table \ref{table:using original images} and Table \ref{table:using images processed by face re-detection and re-alignment} give the comparison of using original images aligned by official\cite{guo2016msceleb,facelowshot} and preprocessed images obtained by our face re-detection and re-alignment. It proves the efficiency of the proposed preprocess scheme.

In addition, Table \ref{table:using original images} and Table \ref{table:using images processed by face re-detection and re-alignment} gives the comparisons of using three different similarity search strategies including exact search for L2, inverted file with exact post-verification and coarse quantizer+PQ on residuals, which corresponding to cpu\_search\_0, cpu\_search\_1 and cpu\_search\_2 respectively. The meaning of ``gpu\_search\_*" is the same as ``cpu\_search\_*", except running on GPUs.

From the results listed in Table \ref{table:using original images} and Table \ref{table:using images processed by face re-detection and re-alignment}, we can conclude that:

1. Exact search for L2 has the highest accuracy. And the program running on GPUs is faster. The speed of search is 1ms/(per image), which is very efficient.

2. Face re-detection and re-alignment boosts the accuracy significantly.

3. For base search set, there are 21,000 individuals. However, do clustering for base search set makes the performance decreased, it seems that the clustering in Faiss\cite{Faiss} for the base search set may not that accurate, which leads to worse results. Other possible reason is the number of images in base search set is not enough for accurate clustering. How to do clustering on base search set to raise the accuracy of face search is our future work.

Based on the results given in Table \ref{table:using original images} and Table \ref{table:using images processed by face re-detection and re-alignment}, we use exact search for L2 on GPUs for face search in challenge 2 of MS-Celeb-1M\cite{guo2016msceleb,facelowshot}. And the speed of searching is 1ms/image, which is very efficient.

\begin{table}
\begin{center}
\scalebox{0.5}
{
\begin{tabular}{|c|c|c|c|}
\hline
search strategy(ori image)&num of clustering center&accuracy on validation set(\%)&time(s)\\
\hline
\hline
cpu\_search\_0&-&72.3&1121.37\\
\hline
cpu\_search\_1&1000&60.6&638.56\\
\hline
cpu\_search\_1&21,000&53.66&4613.36\\
\hline
cpu\_search\_2&1000&32.74&704.06\\
\hline
cpu\_search\_2&21,000&39.56&4678.33\\
\hline
\textbf{gpu\_search\_0}&-&\textbf{72.3}&\textbf{260.51}\\
\hline
gpu\_search\_1&1000&46.7&1121.53\\
\hline
gpu\_search\_1&21,000&-&-\\
\hline
\end{tabular}
}
\end{center}
\caption{Accuracy comparisons of different search strategies(use original images).}\label{table:using original images}
\end{table}

\begin{table}
\begin{center}
\scalebox{0.5}
{
\begin{tabular}{|c|c|c|c|}
\hline
search strategy(re-norm image)&num of clustering center&accuracy on validation set(\%)&time(s)\\
\hline
\hline
cpu\_search\_0&-&86.18&1036.69\\
\hline
cpu\_search\_1&1,000&78.32&279.01\\
\hline
cpu\_search\_1&21,000&70.38&4298.65\\
\hline
cpu\_search\_2&1,000&41.76&367.24\\
\hline
cpu\_search\_2&21,000&41.1&4430.04\\
\hline
\textbf{gpu\_search\_0}&-&\textbf{86.18}&\textbf{255.63}\\
\hline
gpu\_search\_1&1,000&-&-\\
\hline
gpu\_search\_1&21,000&-&-\\
\hline
\end{tabular}
}
\end{center}
\caption{Accuracy comparisons of different search strategies(use images processed by face re-detection and re-alignment).}\label{table:using images processed by face re-detection and re-alignment}
\end{table}

\section{Conclusion}

In this paper, we design a new pipeline for face recogtion/search on large scale datasets. We use image classification and clustering to remove non-face images and error-labeled face images in the base search set. Face re-detection and re-alignment is introduced to make the features more discriminative. And similarity search method improves accuracy and efficiency of face search. Finally, we rank \textbf{3rd} on challenge 2 of MS-Celeb-1M.

{\small
\bibliographystyle{ieee}
\bibliography{egbib}
}

\end{document}